\documentclass{esannV2}
\usepackage[]{hyperref}
\usepackage{graphicx}
\usepackage[utf8]{inputenc}
\usepackage{amssymb,amsmath,array}
\usepackage{bm}
\usepackage{dsfont}
\usepackage{booktabs}

\newcommand{\Var}{\mathrm{Var}}

\begin{document}
\title{Scalable Linearized Laplace Approximation via Surrogate Neural Kernel}

\author{Luis A. Ortega$^{1,3}$, Simón Rodr\'iguez-Santana$^2$ and Daniel Hern\'andez-Lobato$^{1,4}$
%
\thanks{We acknowledge financial support from project PID2022-139856NB-I00, funded by MCIN and from project IDEA-CM (TEC-2024/COM-89) 
and from the ELLIS Unit Madrid, funded by the Autonomous Community of Madrid.  We acknowledge support from Centro de Computaci\'on Cient\'ifica-Universidad Aut\'onoma de Madrid (CCC-UAM).
}
%
\vspace{.3cm}\\
%
1- Machine Learning Group - Computer Science Department \\
Escuela Polit\'ecnica Superior, Universidad Aut\'onoma de Madrid, Spain\vspace{.1cm}\\
2- ICAI Engineering School, Universidad Pontificia de Comillas, Spain\vspace{0.1cm}\\
3- Department of Computer Science, Aalborg University, Copenhagen\vspace{0.1cm}\\
4- Centro de Investigaci\'on Avanzada en F\'isica Fundamental, \\Universidad Aut\'onoma de Madrid, Spain
}

\maketitle

\begin{abstract}
We introduce a scalable method to approximate the kernel of the Linearized Laplace Approximation (LLA). For this, we use a surrogate deep 
neural network (DNN) that learns a compact feature representation whose inner product replicates the Neural Tangent Kernel (NTK). This avoids the 
need to compute large Jacobians.
Training relies solely on efficient Jacobian–vector products, allowing to compute predictive uncertainty on large-scale pre-trained DNNs. 
Experimental results show similar or improved uncertainty estimation and calibration compared to existing LLA approximations. 
Notwithstanding, biasing the learned kernel significantly enhances out-of-distribution detection. This remarks the benefits of the 
proposed method for finding better kernels than the NTK in the context of LLA to compute prediction uncertainty given a pre-trained DNN.
\end{abstract}

\section{Introduction}
Bayesian methods offer a principled framework for uncertainty estimation in deep neural networks (DNNs), providing calibrated predictions and robustness to distributional shifts \cite{mackay1992}. However, their scalability remains limited by the high computational cost of posterior inference. 
The \textit{Linearized Laplace Approximation (LLA)} \cite{immer2021} provides a practical \textit{post-hoc} Bayesian approach by linearizing the DNN around its trained parameters, yielding a Gaussian posterior approximation governed by the \textit{Neural Tangent Kernel (NTK)} \cite{jacot2018}. 
Despite its appeal, the need to compute and store large Jacobians constrains its use in modern DNNs.

We propose the use of a \textit{surrogate neural network} that learns to approximate the NTK without explicit Jacobian computation.  
The model generates compact feature embeddings whose inner products emulate the NTK. Such a model is trained efficiently via \textit{Jacobian–vector products (JVPs)}.  
This enables a scalable, memory-efficient approximation to LLA suitable for large DNN architectures.

Our main contributions are (i) a \textit{Jacobian-free LLA approximation} using a learned surrogate kernel with; (ii) improved \textit{scalability and calibration} compared to existing LLA variants, and (iii) a mechanism to \textit{bias the learned covariances} resulting in enhanced out-of-distribution data detection (OOD).

Overall, the proposed approach bridges Bayesian posterior uncertainty and scalability, allowing for LLA-like estimates in large DNNs.

\section{Methodology}

The goal of the proposed method is to approximate the kernel used in the LLA without explicitly computing the network Jacobian, an expensive task since it contains as many elements as weights has the given pre-trained DNN. We achieve this by training a \textit{surrogate neural network} to emulate the behavior of the NTK, enabling scalable Bayesian inference for large neural architectures.

\subsection{Background: The Linearized Laplace Approximation}

The Linearized Laplace Approximation (LLA) \cite{immer2021} models a trained neural network \( f_{\bm{\theta}}(\mathbf{x}) \), with weights $\bm{\theta}$, as a first-order Taylor expansion around the maximum a posteriori (MAP) parameters \( \bm{\theta}^\ast \), found, \emph{e.g.}, via back-propagation: 
\begin{equation}
    f_{\bm{\theta}}(\mathbf{x}) \approx f_{\bm{\theta}^\ast}(\mathbf{x}) 
    + J_{\bm{\theta}^\ast}(\mathbf{x})(\bm{\theta} - \bm{\theta}^\ast),
\end{equation}
where \( J_{\bm{\theta}^\ast}(\mathbf{x}) := \nabla_{\bm{\theta}}f_{\bm{\theta}^\ast}(\mathbf x) \in \mathds{R}^{C \times P} \) is the Jacobian of the network outputs with respect to the parameters \( \bm{\theta} \), evaluated at \( \bm{\theta}^\ast \).  
Here, \( C \) denotes the number of output dimensions and \( P \) the number of parameters.

Under a Gaussian prior on the weights \( \bm{\theta} \sim \mathcal{N}(0, \sigma_0^2\mathbf{I}) \), LLA defines a Gaussian posterior approximation of the DNN weights, with covariance given by the inverse of the \textit{Generalized Gauss–Newton} (GGN) precision matrix:
\begin{align}
    \bm{\Sigma}_{\text{GGN}}^{-1}
	& = \textstyle \sum_{n=1}^{N}
      J_{\bm{\theta}^\ast}(\mathbf{x}_n)^{\!\top}
      \, \bm{\Lambda}(\mathbf{y}_n; \mathbf{f}_n)
      \, J_{\bm{\theta}^\ast}(\mathbf{x}_n)
      + \sigma_0^{-2}\mathbf{I},
    \label{eq:ggn_precision}
\end{align}
where \( \bm{\Lambda}(\mathbf{y}_n; \mathbf{f}_n) := -\nabla_{\mathbf {ff}}^2 \log p(\mathbf y_n | \mathbf f_n)\) is the local curvature of the negative log-likelihood with respect to the DNN outputs \( \mathbf{f}_n := f_{\bm{\theta}^\ast}(\mathbf{x}_n) \).  
In classification, $p(\mathbf y_n | \mathbf f_n)$ is simply given by the soft-max activation function.

Eq.~(\ref{eq:ggn_precision}) shows that \emph{learning} in LLA does not involve further optimization of \( \bm{\theta} \), but rather a single pass over the training data to accumulate the curvature contributions
\(
    J_{\bm{\theta}^\ast}(\mathbf{x}_n)^{\!\top}
    \bm{\Lambda}(\mathbf{y}_n; \mathbf{f}_n)
    J_{\bm{\theta}^\ast}(\mathbf{x}_n)
\)
, \( \forall \, n \in \{1, \ldots, N\} \).
This accumulation produces the auxiliary precision matrix \( \bm{\Sigma}_{\text{GGN}}^{-1} \), which is then used for posterior inference. Once this matrix is available, the predictive distribution for a new input \( \mathbf{x}_\ast \) is given by \(\mathcal{N}(f_{\bm{\theta}^\ast}(\mathbf{x}^\ast), K_{\mathrm{LLA}}(\mathbf{x}^\ast, \mathbf{x}^\ast))\), where
\(
    K_{\mathrm{LLA}}(\mathbf{x}_1, \mathbf{x}_2) := J_{\bm{\theta}^\ast}(\mathbf{x}_1)
           \bm{\Sigma}_{\text{GGN}}
           J_{\bm{\theta}^\ast}(\mathbf{x}_2)^{\top} \label{eq:lla_pred_cov}
\) is the LLA kernel or posterior covariance. 

Although \eqref{eq:ggn_precision} does not involve gradient descent, it requires iterating over the entire training data to accumulate curvature contributions. 
This demands full Jacobian evaluations for all training points, resulting in expensive computations for many modern DNN architectures and datasets. 
Inverting $\bm{\Sigma}_{\text{GGN}}^{-1}$ has a $\mathcal{O}(P^3)$ computational cost, which can be reduced to $\mathcal{O}(N^3)$ if the Woodbury matrix inversion lemma is used to compute $ J_{\bm{\theta}^\ast}(\mathbf{x}_1)
\bm{\Sigma}_{\text{GGN}} J_{\bm{\theta}^\ast}(\mathbf{x}_2)^{\top}$ \cite{immer2021}.

Importantly, LLA predictive variances can be expressed directly in terms of the NTK, defined as \( K_{\mathrm{NTK}}(\mathbf{x}, \mathbf{x}') := J_{\bm{\theta}^\ast}(\mathbf{x}) J_{\bm{\theta}^\ast}(\mathbf{x}')^{\top} \) \cite{immer2021}.
This relationship implies that the entire LLA kernel structure depends on the inner products of Jacobian features captured by the NTK.  
In other words, learning or approximating the LLA kernel is equivalent to learning how the NTK behaves across the data manifold.  
Therefore, instead of explicitly computing all Jacobians to form \( K_{\mathrm{LLA}} \), one can learn a surrogate model 
that reproduces the NTK structure in feature space. By doing so, the LLA kernel (and consequently, the predictive posterior) can be 
recovered implicitly, enabling scalable Laplace inference without forming or storing large Jacobian matrices \cite{immer2021}.

\subsection{Surrogate Kernel Approximation}

We introduce a surrogate DNN, denoted \({g_{\bm{\phi}}(\mathbf{x})}: \mathcal{X} \rightarrow \mathds{R}^{C \times m}\), which learns a mapping with \(m \ll P\), such that the inner product \(\tilde{K}(\mathbf{x}, \mathbf{x}') = g_{\bm{\phi}}(\mathbf{x}) g_{\bm{\phi}}(\mathbf{x}')^{\top}\) approximates the NTK kernel \( K_{\mathrm{NTK}}(\mathbf{x}, \mathbf{x}') \).  
This reduces the need to compute \( J_{\bm{\theta}^\ast} \) explicitly, as \( g_{\bm{\phi}}(\mathbf{x}) \) acts as a compressed representation of the Jacobian features.

Computing or storing the full Jacobian matrix \( J_{\bm{\theta}^\ast}(\mathbf{x}) \in \mathds{R}^{C \times P} \) is intractable for large models, as it requires \(\mathcal{O}(CP)\) memory. 
For that reason, the proposed approach to fit the surrogate model relies on efficient Jacobian–vector product (JVP) computations, which can be obtained efficiently through automatic differentiation without forming the full Jacobian \cite{deng2022accelerated}. 

The surrogate model is trained to match these projected outputs by minimizing the following loss
\begin{align}
	\mathcal{L}(\bm{\phi}) & = \mathds{E}_{\mathbf{x}, \mathbf{x}',
    \mathbf{v}}\left[\| g_{\bm{\phi}}(\mathbf{x}) g_{\bm{\phi}}(\mathbf{x}')^{\top} - J_{\bm{\theta}^\ast}(\mathbf{x})\mathbf{v} (J_{\bm{\theta}^\ast}(\mathbf{x}')\mathbf{v})^{\top} \|^2 \right],
	\label{eq:loss_proposed}
\end{align}
which encourages \( \tilde{K}(\mathbf{x}, \mathbf{x}') \) to approximate \( K_{\mathrm{NTK}}(\mathbf{x}, \mathbf{x}') \) in expectation over random projections \( \mathbf{v} \).  
This training scheme preserves the kernel’s local geometry while avoiding the explicit computation or storage of the network's Jacobians.
Note that Eq. (\ref{eq:loss_proposed}) allows for efficient mini-batch optimization methods.

For any random vector \( \mathbf{v} \in \mathds{R}^{P\times 1} \) with zero mean and unit covariance, the following unbiased identity holds:
\begin{align}
    \mathds{E}_{\mathbf{v}}
	\big[ J_{\bm{\theta}^\ast}(\mathbf{x})\mathbf{v} (J_{\bm{\theta}^\ast}(\mathbf{x}')\mathbf{v})^{\top} \big]
    = \mathds{E}_{\mathbf{v}}
	\big[ J_{\bm{\theta}^\ast}(\mathbf{x})\mathbf{v} \mathbf{v}^{\top} J_{\bm{\theta}^\ast}(\mathbf{x}')^{\top}\big]
    = J_{\bm{\theta}^\ast}(\mathbf{x})J_{\bm{\theta}^\ast}(\mathbf{x}')^\top .
    \label{eq:trace_estimator}
\end{align}
Hence, the NTK can be approximated by JVPs, 
\(
    \mathbf{z}_{\mathbf{v}}(\mathbf{x}) := J_{\bm{\theta}^\ast}(\mathbf{x}) \mathbf{v}, 
\)
and taking their inner product
\(
    \mathbf{z}_{\mathbf{v}}(\mathbf{x}) \mathbf{z}_{\mathbf{v}}(\mathbf{x}')^\top
\).
This estimator is unbiased and can be computed without the explicit \(J_{\bm{\theta}^\ast}\).  For \(\mathbf{v} \sim \mathcal{N}(\mathbf{0}, \mathbf{I})\),
\begin{align}
	\Var_{\text{Gauss}}\left( \mathbf{z}_{\mathbf{v}}(\mathbf{x}) \mathbf{z}_{\mathbf{v}}(\mathbf{x})^\top\right) = 2\,\mathrm{tr}\left(J_{\bm{\theta}^\ast}(\mathbf x)^\top J_{\bm{\theta}^\ast}(\mathbf x)\right)^2,
\end{align}
which grows with the norm of \(J_{\bm{\theta}^\ast}\).  When \(\mathbf{v}\) is drawn from a Rademacher distribution (\(v_i \in \{-1, 1\}\) uniformly), we have
\(
\mathds{E}[\mathbf{v} \mathbf{v}^\top] = \mathbf{I}
\),
but the fourth moment is smaller than for a Gaussian.  
Consequently, the variance becomes
\begin{align}
	\Var_{\mathrm{Rad}}\left( \mathbf{z}_{\mathbf{v}}(\mathbf{x}) \mathbf{z}_{\mathbf{v}}(\mathbf{x})^\top\right) = 2\,\mathrm{tr}\left(J_{\bm{\theta}^\ast}(\mathbf x)^\top J_{\bm{\theta}^\ast}(\mathbf x)\right)^2 - 2 \| \mathrm{diag}(J_{\bm{\theta}^\ast}(\mathbf x)^{\top} J_{\bm{\theta}^\ast}(\mathbf x))\|_2^2 .
\end{align}
Therefore, Rademacher projections show a slightly lower-variance estimator than Gaussian ones. Hence, we employ them in our method.

To ensure that the surrogate model generalizes beyond the training data, the learned feature representation exhibits a Jacobian structure similar to that of the original model, not only on the training set but also within nearby regions of the input space. 
To achieve this, we introduce a collection of \textit{context points}, sampled independently from an auxiliary dataset that resembles the evaluation data, potentially coming from a distribution different from that of the training set. During training, the input to the surrogate model $g_\phi(\cdot)$ is formed by concatenating a batch of training samples with a batch of context points:
\begin{align}
	\mathbf{X}_{\text{batch}} \, = \,  &
    [\, \mathbf{X}_{\text{train}}^{(b)} \,;\, \mathbf{X}_{\text{context}}^{(b)} \,].
\end{align}
The goal is to ensure that the surrogate kernel \( g_{\bm{\phi}}(\mathbf{x})  g_{\bm{\phi}}(\mathbf{x}')^\top \) closely matches the NTK kernel \( J_{\bm{\theta}^\ast}(\mathbf{x}) J_{\bm{\theta}^\ast}(\mathbf{x})^\top \) over the extended set. This enforces the surrogate model to approximate the LLA predictive variances across both in-distribution and OOD regions. Given $\mathbf{X}_{\text{batch}}$, and the surrogate features, the empirical kernels
\begin{align}
	K_{ijab} & = \mathbf{z}_{\mathbf{v}}(\mathbf{x}_i) \mathbf{z}_{\mathbf{v}}(\mathbf{x}_j)^\top, &
	Q_{ijab} & = g_{\bm{\phi}}(\mathbf{x}_i) g_{\bm{\phi}}(\mathbf{x}_j)^\top,
\end{align}
are computed in batched form, where \(a\) and \(b\) denote the output dimension (class labels). We then
perform a gradient step to minimize $||\mathbf{K} - \mathbf{Q}||^2$. 

\subsection{Biasing the Learned Kernel}

To enhance OOD sensitivity, we intentionally bias the covariance structure by zeroing the cross-covariances between the training and context points:
\begin{align}
	K_{ijab} & = 0 \quad \text{if} \quad 
    \mathbf{x}_i \in \mathbf{X}_{\text{train}}, \,
    \mathbf{x}_j \in \mathbf{X}_{\text{context}}
    \;\text{or vice versa.}
\end{align}
This operation effectively enforces a block-diagonal structure in the covariance matrix given by the approximate kernel, indicating independence, and emphasizing that training and context regions belong to different input distributions.

By suppressing correlations between in-distribution and context points, the resulting kernel is expected to enforce a stronger back-to-the-prior behavior between known 
and unknown regions of the input space. This bias improves the model’s ability to detect OOD samples: predictions for inputs far from the training manifold are 
expected to yield higher posterior variance, while maintaining accurate calibration on in-distribution data. 
Empirically, this covariance biasing significantly enhances OOD detection without requiring additional regularization or auxiliary loss functions.

By avoiding explicit Jacobian storage and relying solely on JVPs computations and mini-batch training, the proposed approach achieves linear scalability with respect to the number of model parameters and sub-linear with respect to training samples.  
The resulting kernel representation \( g_{\bm{\phi}}(\mathbf{x}) \) is compact and reusable across inference tasks, enabling scalable inference even for DNNs with millions of parameters and large datasets. 

\subsection{Limitations}

A key limitation arises when enhancing OOD detection through biased covariance learning.
The effectiveness of this bias strongly depends on the choice of \textit{context points}, \emph{i.e.}, the data used to condition the surrogate model.  
To achieve optimal OOD sensitivity, these context points must be representative of the out-of-distribution region against which OOD samples will be compared. 
Notwithstanding, since OOD samples are not available during training, finding a suitable dataset of context points can be challenging in practice.

\section{Experimentation}

We employ a convolutional neural network (CNN) trained on the FMNIST dataset, with the model weights given by the back-propagation estimate \( \bm{\theta}^\ast \). 
In our proposed method, scalable approximate LLA (\textbf{ScaLLA}), we train the surrogate network \( g_{\bm{\phi}}(\mathbf{x}) \) using JVPs derived from the CNN, as described in the previous section. 
This network is equivalent to the initial CNN but with $100$ outputs per class, \emph{i.e.}, $m=100$.
The context dataset is the MNIST test set. 
For OOD evaluation we follow the standard setup, treating the test set of KMNIST as OOD and the test set of FMNIST as in-distribution.
All hyper-parameters are tuned by maximizing the LLA estimate of the marginal likelihood \cite{immer2021}. 
The code to reproduce all experiments is at \url{http://www.github.com/Ludvins/BayesiPy}.

We report OOD detection performance in terms of AUC-ROC, where the entropy of the predictive distribution (after computing soft-max probabilities) is used for discrimination \cite{ortega2024,ortega2025}.
Additionally, we report results when the entropy of the Gaussian predictive distribution is used (if the method outputs a 
Gaussian distribution). The quality of the predictive distribution is evaluated using the negative log-likelihood (NLL) and the expected
calibration error (ECE) on the test set, following \cite{ortega2024}.

ScaLLA is compared against state-of-the-art Bayesian approximations and uncertainty-aware deep learning methods: (i) Last-layer KFAC Linearized Laplace Approximation (\textbf{LLLA}) \cite{immer2021}; 
(ii) a variational Laplace approximation over the linearized model (\textbf{VaLLA}) \cite{ortega2024};  
(iii) fixed-Mean Gaussian Processes (\textbf{FMGP}) \cite{ortega2025}, a scalable GP interpretation of deep feature posteriors;
(iv) mean-field variational inference, a baseline for standard variational Bayesian neural networks (\textbf{MFVI}) \cite{deng2022accelerated};
 and (v) spectral-normalized Gaussian Processes (\textbf{SNGP}) \cite{liu2023simple}, which enhance uncertainty estimation via spectral normalization and GP regression layers. Note that LLA and LLA-KFAC are intractable in this context \cite{immer2021}.

Table~\ref{tab:results} summarizes the quantitative results across all methods. 
We observe that ScaLLA maintains a good predictive distribution for test data in terms of NLL and ECE.
Moreover, by biassing the learnt covariance structure, the method achieves superior OOD detection without 
sacrificing accuracy or calibration.
LLLA also achieves the best OOD detection but degrades the predictions for in-distribution data.
These findings suggest that learning a surrogate representation of the LLA kernel offers an effective trade-off between scalability, uncertainty calibration, and OOD detection performance.

\begin{table}[t!]
\centering
\caption{Comparison of predictive performance and uncertainty metrics.}
\label{tab:results}
\scalebox{1}{
\begin{tabular}{lccccc}
\toprule
\textbf{Method} & \textbf{ACC} $\uparrow$ & \textbf{NLL} $\downarrow$ & \textbf{ECE} $\downarrow$ & \textbf{AUC} $\uparrow$ & \textbf{Gaus. AUC} $\uparrow$ \\
\midrule
MAP &  $91.01$  &  $0.2539$  &  $0.010$  &  $0.773$ & -- \\
LLA &  $90.92$  &  $0.2714$  &  $0.023$  &  $0.841$  &  $\textbf{0.992}$  \\
VaLLA &  $90.98$  &  $\textbf{0.2516}$  &  $\textbf{0.005}$  &  $0.677$  &  $0.939$  \\
FMGP &  $91.00$  &  $0.2623$  &  $0.036$  &  $\textbf{0.890}$  &  $0.866$  \\
MFVI &  $\textbf{91.06}$  &  $0.2541$  &  $0.012$  &  $0.783$ & -- \\
SNGP &   $90.36$  &  $0.2674$  &  $0.011$  &  $\textbf{0.927}$  & -- \\
\textbf{ScaLLA} &  $\textbf{91.03}$  &  $\textbf{0.2528}$  &  $\textbf{0.008}$  &  $0.789$  &  $0.784$  \\
\textbf{ScaLLA biased} &  $90.99$  &  $0.2530$  &  $\textbf{0.008}$  &  $0.821$  &  $\textbf{0.982}$  \\
\bottomrule
\end{tabular}}
\end{table}

\section{Conclusions}

This work introduced ScaLLA, a scalable method to approximate the kernel of the Linearized Laplace Approximation (LLA) using a surrogate neural network.  
By exploiting efficient Jacobian–vector products to learn compact feature representations, ScaLLA removes the need for explicit Jacobian computation and scales effectively to large datasets and DNNs. 
Experiments show that ScaLLA matches or outperforms existing LLA variants and other methods in both accuracy and calibration for in-distribution data, while achieving notable improvements in out-of-distribution detection.
These results highlight the promise of learned kernel surrogates for extending Bayesian uncertainty estimation to large-scale neural networks.


\begin{footnotesize}




\bibliographystyle{unsrt}
\bibliography{references}

\end{footnotesize}


\end{document}